%% file: NN_ImgProc.tex
\documentclass[journal]{IEEEtran}

\usepackage{times}
\usepackage{epsfig}
\usepackage{graphicx}
\usepackage{amsmath}
\usepackage{amssymb}
\usepackage{color}
\usepackage{subfig}
\usepackage{xspace}
\usepackage{comment}
\usepackage[normalem]{ulem}
\usepackage{lineno}
\usepackage{wrapfig}
\usepackage{afterpage}

\usepackage{rotating}
\usepackage[online]{threeparttable}
\usepackage{multirow}

\newcommand{\lone}{$\ell_1$\xspace}
\newcommand{\ltwo}{$\ell_2$\xspace}

\definecolor{darkblue}{RGB}{0,0,175}

\usepackage[pagebackref=true,breaklinks=true,letterpaper=true,colorlinks,bookmarks=false]{hyperref}

\ifCLASSINFOpdf
\else
\fi

\begin{document}

\title{Loss Functions for Image Restoration with Neural Networks}

\author{Hang Zhao$^{\star,\dagger}$, Orazio Gallo$^\star$, Iuri Frosio$^\star$, and Jan Kautz$^\star$
\thanks{$^\star$NVIDIA,~~$^\dagger$MIT Media Lab

 The supplementary material is available at \url{http://research.nvidia.com/publication/loss-functions-image-restoration-neural-networks}. The material includes additional numerical and visual comparisons, and mathematical derivations. Contact ogallo@nvidia.com for further questions about this work.}%
}

\markboth{}%
{}

\maketitle

\begin{abstract}
\input{abstract}
\end{abstract}


\IEEEpeerreviewmaketitle


\section{Introduction}\label{sec:intro}
\input{intro}

\section{Related work}\label{sec:related}
\input{related}

\section{Loss layers for image restoration}\label{sec:layers}
\input{new_loss_layers}

\section{Results}\label{sec:results}
\input{results}

\input{discussion}

\section{Conclusions}
\input{conclusions}

\bibliographystyle{splncs}
\bibliography{biblio}

\begin{IEEEbiography}[{\includegraphics[width=1in,height=1.25in,clip,keepaspectratio]{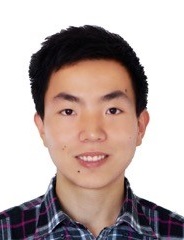}}]{Hang Zhao}
received a B.E. degree in information engineering from Zhejiang University, China in 2013. He is currently a Ph.D. student at the Massachusetts Institute of Technology. His research interests include computer vision, machine learning, and their robotic applications.
\end{IEEEbiography}

\begin{IEEEbiography}[{\includegraphics[width=1in,height=1.25in,clip,keepaspectratio]{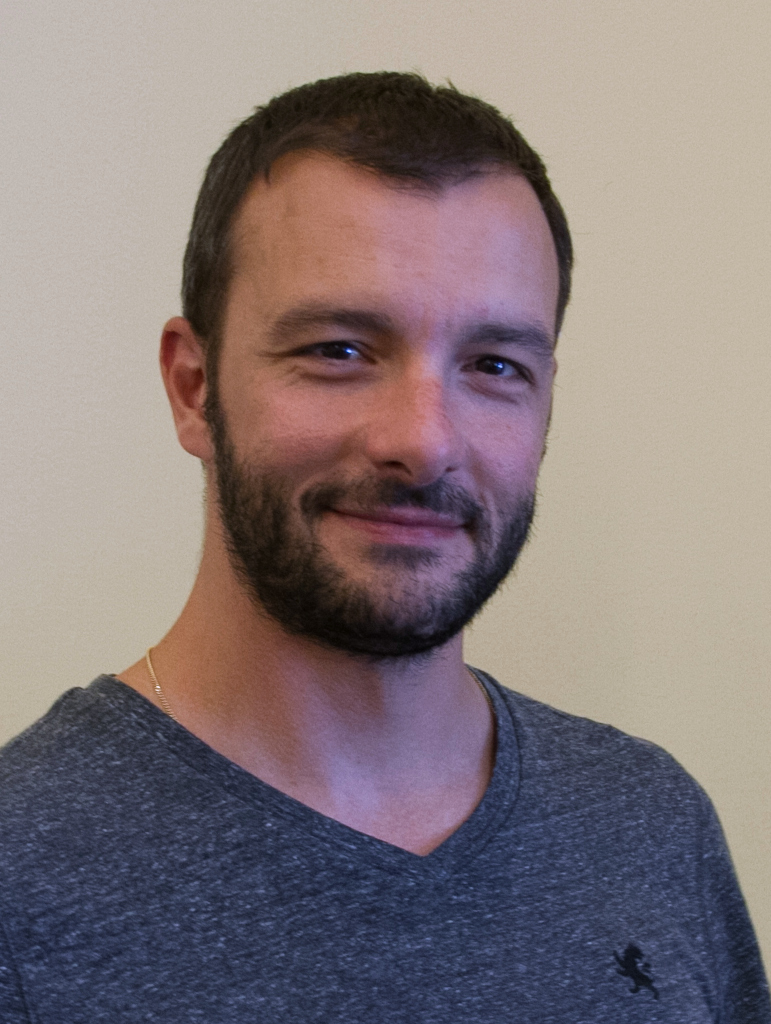}}]{Orazio Gallo}
earned a M.S. degree in Biomedical Engineering from Politecnico di Milano in 2004, and a Ph.D. in Computer Engineering from the University of California at Santa Cruz in 2011. From 2004 to 2006 he was a research assistant at the Smith-Kettlewell Eye Research Institute. In 2011, Orazio joined NVIDIA, where he is a senior research scientist. His research interests range from computational photography, to image processing,  applied visual perception, and computer vision.
Orazio is also an associate editor of the IEEE Transactions of Computational Imaging, as well as Signal Processing: Image Communication.
\end{IEEEbiography}

\begin{IEEEbiography}[{\includegraphics[width=1in,height=1.25in,clip,keepaspectratio]{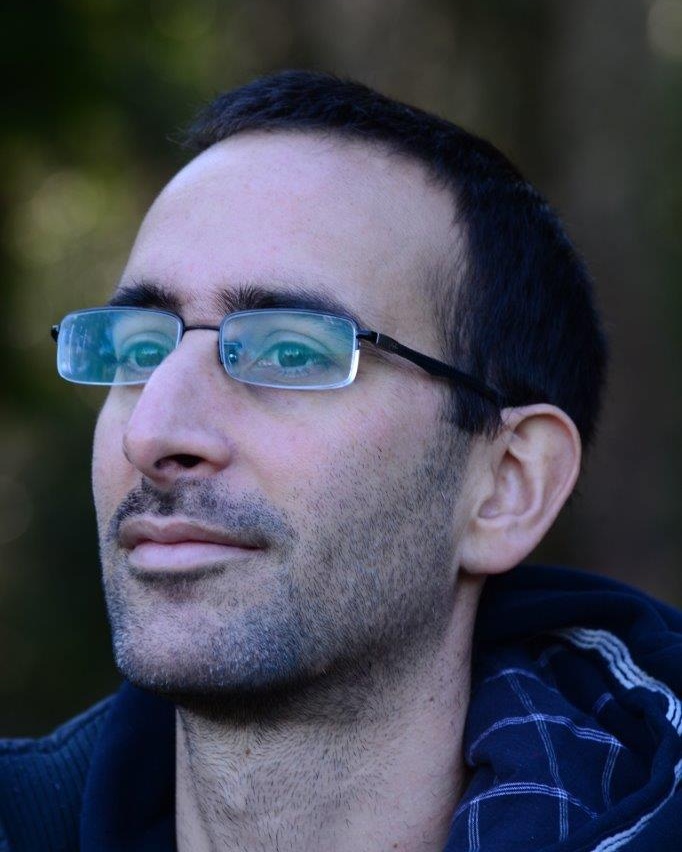}}]{Iuri Frosio}
got his M.S. and Ph.D. degrees in biomedical engineering at the Politecnico di Milano in 2003 and 2006, respectively. He was a research fellow at the Computer Science Department of the Universit\'{a} degli Studi di Milano from 2003 to 2006, and an assistant professor in the same Department from 2006 to 2013. In the same period, he worked as a consultant for various companies in Italy and abroad. He joined NVIDIA in 2014 as senior research scientist. His research interests include image processing, computer vision, inertial sensors, and machine learning. Iuri is also an associate editor of the Journal of Electronic Imaging.
\end{IEEEbiography}

\begin{IEEEbiography}[{\includegraphics[width=1in,height=1.25in,clip,keepaspectratio]{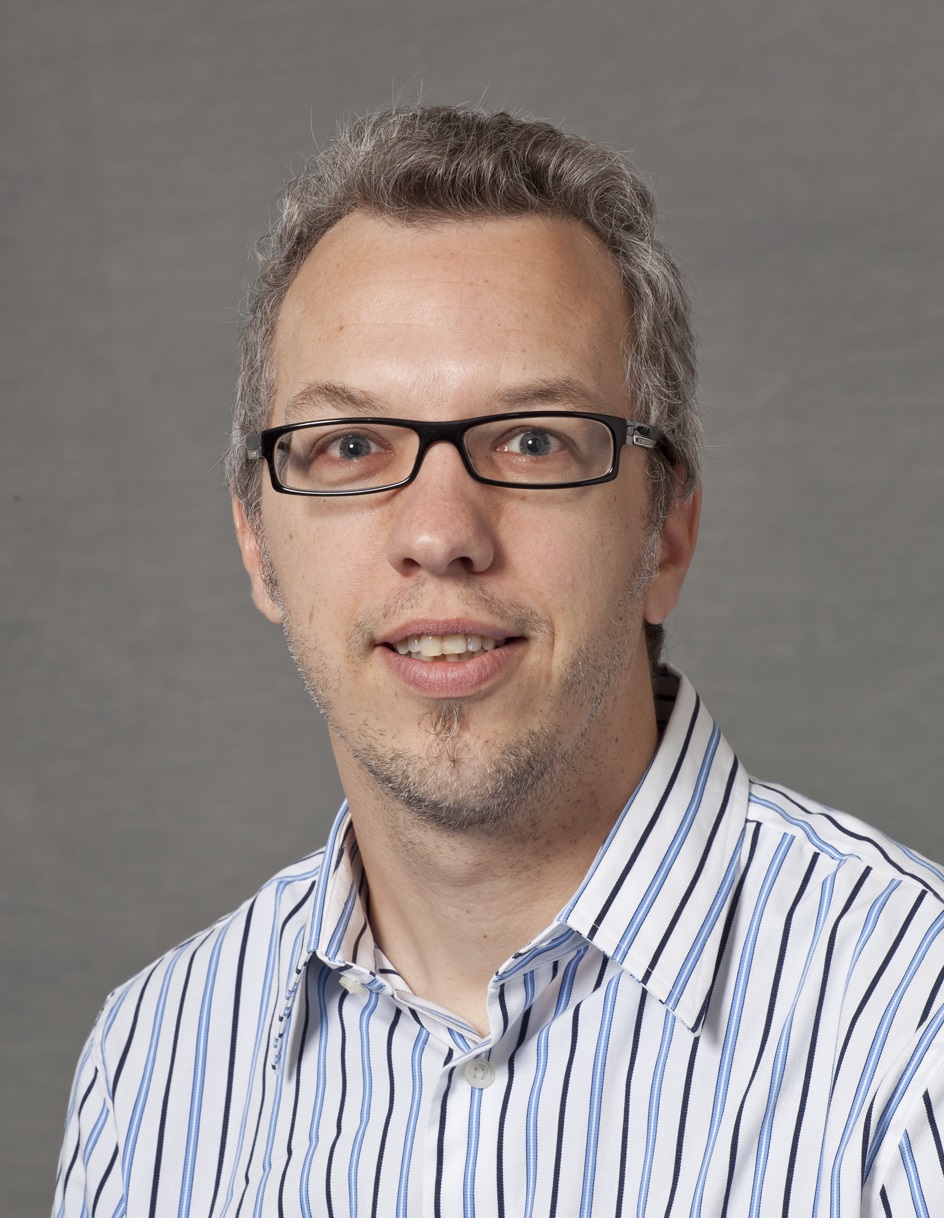}}]{Jan Kautz}
leads the Visual Computing Research team at NVIDIA, working predominantly on computer vision problems---from low-level vision (denoising, super-resolution, computational photography) and geometric vision (structure from motion, SLAM, optical flow) to high-level vision (detection, recognition, classification), as well as machine learning problems such as deep reinforcement learning and efficient deep learning. Before joining NVIDIA in 2013, Jan was a tenured faculty member at University College London. He holds a BSc in Computer Science from University of Erlangen-N{\"u}rnberg (1999), an MMath from the University of Waterloo (1999), received his Ph.D. from the Max-Planck-Institut f{\"u}r Informatik (2003), and worked as a post-doc at the Massachusetts Institute of Technology (2003-2006). 

Jan was program co-chair of the Eurographics Symposium on Rendering 2007, program chair of the IEEE Symposium on Interactive Ray-Tracing 2008, program co-chair for Pacific Graphics 2011, and program chair of CVMP 2012. He co-chaired Eurographics 2014 and was on the editorial board of IEEE Transactions on Visualization \& Computer Graphics as well as The Visual Computer.
\end{IEEEbiography}


\end{document}

%% file: abstract.tex
Neural networks are becoming central in several areas of computer vision and image processing and different architectures have been proposed to solve specific problems. The impact of the loss layer of neural networks, however, has not received much attention in the context of image processing: the default and virtually only choice is \ltwo. In this paper, we bring attention to alternative choices for image restoration. In particular, we show the importance of perceptually-motivated losses when the resulting image is to be evaluated by a human observer. We compare the performance of several losses, and propose a novel, differentiable error function. We show that the quality of the results improves significantly with better loss functions, even when the network architecture is left unchanged.

%% file: intro.tex

\IEEEPARstart{F}{or} decades, neural networks have shown various degrees of success in several fields, ranging from robotics, to regression analysis, to pattern recognition. 
Despite the promising results already produced in the 1980s on handwritten digit recognition~\cite{Lecun89}, the popularity of neural networks in the field of computer vision has grown exponentially only recently, when deep learning boosted their performance in image recognition~\cite{Krizhevsky12}.

In the span of just a couple of years, neural networks have been employed for virtually every computer vision and image processing task known to the research community. Much research has focused on the definition of new architectures that are better suited to a specific problem~\cite{Burger12,xu2014}.
A large effort was also made to understand the inner mechanisms of neural networks, and what their intrinsic limitations are, for instance through the development of deconvolutional networks~\cite{Zeiler10}, or trying to fool networks with specific inputs~\cite{nguyen15}.
Other advances were made on the techniques to improve the network's convergence~\cite{Hinton12}.

The loss layer, despite being the effective driver of the network's learning, has attracted little attention within the image processing research community: the choice of the cost function generally defaults to the squared \ltwo norm of the error~\cite{Jain09,Burger12,Fleet14,Wang14}. 
This is understandable, given the many desirable properties this norm possesses.
There is also a less well-founded, but just as relevant reason for the continued popularity of \ltwo: standard neural networks packages, such as Caffe~\cite{caffe}, only offer the implementation for this metric.

However, \ltwo suffers from well-known limitations.
For instance, when the task at hand involves image quality, \ltwo correlates poorly with image quality as perceived by a human observer~\cite{Zhang12}. This is because of a number of assumptions implicitly made when using \ltwo.
First and foremost, the use of \ltwo assumes that the impact of noise is independent of the local characteristics of the image. On the contrary, the sensitivity of the Human Visual System (HVS) to noise depends on local luminance, contrast, and structure~\cite{SSIM}.
The \ltwo loss also works under the assumption of white Gaussian noise, which is not valid in general.

\begin{figure*}[t]
	\includegraphics[width=\textwidth, trim=2cm 19.1cm 2.6cm 3.6cm, clip=true, scale=1]{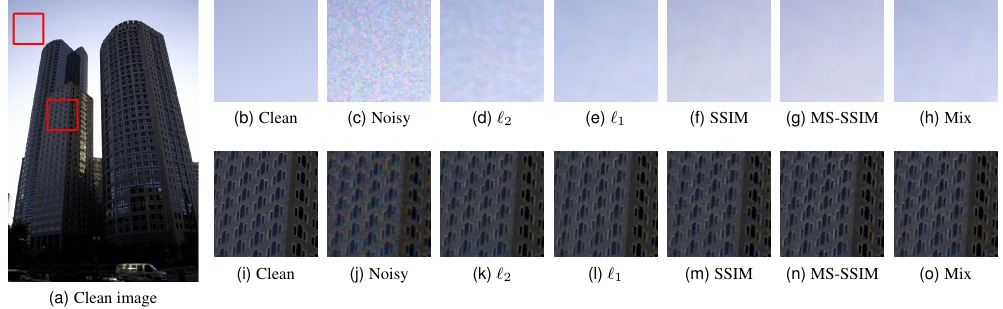}
	\captionof{figure}{Comparisons of the results of joint denoising and demosaicking performed by networks trained on different loss functions (best viewed in the electronic version by zooming in). \ltwo, the standard loss function for neural networks for image processing, produces splotchy artifacts in flat regions (d).}
	\label{fig:teaser}
\end{figure*}

We focus on the use of neural networks for image restoration tasks, and we study the effect of different metrics for the network's loss layer. 
We compare \ltwo against four error metrics on representative tasks: image super-resolution, JPEG artifacts removal, and joint denoising plus demosaicking. First, we test whether a different local metric such as \lone can produce better results. We then evaluate the impact of perceptually-motivated metrics. We use two state-of-the-art metrics for image quality: the structural similarity index (SSIM~\cite{SSIM}) and the multi-scale structural similarity index (MS-SSIM~\cite{MSSSIM}). We choose these among the plethora of existing indexes, because they are established measures, and because they are differentiable---a requirement for the backpropagation stage. As expected, on the use cases we consider, the perceptual metrics outperform \ltwo. However, and perhaps surprisingly, this is also true for \lone, see Figure~\ref{fig:teaser}. 
Inspired by this observation, we propose a novel loss function and show its superior performance in terms of all the metrics we consider.

We offer several contributions. First we bring attention to the importance of the error metric used to train neural networks for image processing: despite the widely-known limitations of \ltwo, this loss is still the \emph{de facto} standard. We investigate the use of three alternative error metrics (\lone, SSIM, and MS-SSIM), and define a new metric that combines the advantages of \lone and MS-SSIM (Section~\ref{sec:layers}). We also perform a thorough analysis of their performance in terms of several image quality indexes (Section~\ref{sec:results}). Finally, we discuss their convergence properties. We empirically show that the poor performance of some losses is related to local minima of the loss functions (Section~\ref{sec:convergence}), and we explain the reasons why SSIM and MS-SSIM alone do not produce the expected quality (Section~\ref{sec:SSIMperf}).
For each of the metrics we analyze, we implement a loss layer for Caffe, which we make available to the research community\footnote{\url{https://github.com/NVlabs/PL4NN}}.

%% file: related.tex

In this paper, we target neural networks for image restoration, which, in our context, is the set of all the image processing algorithms whose goal is to output an image that is appealing to a human observer. To this end, we use the problems of super-re\-so\-lution, JPEG artifacts removal, and joint demosaicking plus denoising as benchmark tests. Specifically, we show how established error measures can be adapted to work within the loss layer of a neural network, and how this can positively influence the results. Here we briefly review the existing literature on the subject of neural networks for image processing, and on the subject of measures of image quality.

\subsection{Neural networks for image restoration}
Following their success in several computer vision tasks~\cite{He15,Krizhevsky12}, neural networks have received considerable attention in the context of image restoration. Neural networks have been used for denoising~\cite{Jain09,Burger12}, deblurring~\cite{xu2014}, demosaicking~\cite{Wang14}, and super-res\-o\-lu\-tion~\cite{Fleet14} among others. To the best of our knowledge, however, the work on this subject has focused on tuning the architecture of the network for the specific application; the loss layer, which effectively drives the learning of the network to produce the desired output quality, is based on \ltwo for all of the approaches above.

We show that a better choice for the error measure has a strong impact on the quality of the results. Moreover, we show that even when \ltwo is the appropriate loss, alternating the training loss function with a related loss, such as \lone, can lead to finding a better solution for \ltwo.

\vspace{-1mm}

\subsection{Evaluating image quality}\label{sec:imQuality}
The mean squared error, \ltwo, is arguably the dominant error measure across very diverse fields, from regression problems, to pattern recognition, to signal and image processing. Among the main reasons for its popularity is the fact that it is convex and differentiable---very convenient properties for optimization problems. Other interesting properties range from the fact that \ltwo provides the maximum likelihood estimate in case of independent and identically distributed Gaussian noise, to the fact that it is additive for independent noise sources. There is longer list of reasons for which we refer the reader to the work of Wang and Bovik~\cite{Wang09}.

These properties paved the way for \ltwo's widespread adoption, which was further fueled by the fact that standard software packages tend to include tools to use \ltwo, but not many other error functions for regression. In the context of image processing, Caffe~\cite{caffe} actually offers \emph{only} \ltwo as a loss layer\footnote{Caffe indeed offers other types of loss layers, but they are only useful for classification tasks.}, thus discouraging researchers from testing other error measures.

However, it is widely accepted that \ltwo, and consequently the Peak Signal-to-Noise Ratio, PSNR, do not correlate well with human's perception of image quality~\cite{Zhang12}: \ltwo simply does not capture the intricate characteristics of the human visual system (HVS).

There exists a rich literature of error measures, both reference-based and non re\-fe\-rence-based, that attempt to address the limitations of the simple \ltwo error function. For our purposes, we focus on reference-based measures.
A popular reference-based index is the structural similarity index (SSIM~\cite{SSIM}). SSIM evaluates images accounting for the fact that the HVS is sensitive to changes in local structure.
Wang et al.~\cite{MSSSIM} extend SSIM observing that the scale at which local structure should be analyzed is a function of factors such as image-to-observer distance. To account for these factors, they propose MS-SSIM, a multi-scale version of SSIM that weighs SSIM computed at different scales according to the sensitivity of the HVS.
Experimental results have shown the superiority of SSIM-based indexes over \ltwo. As a consequence, SSIM has been widely employed as a metric to evaluate image processing algorithms. Moreover, given that it can be used as a differentiable cost function, SSIM has also been used in iterative algorithms designed for image compression~\cite{Wang09}, image reconstruction~\cite{Brunet10}, denoising and super-resolution~\cite{Rehman12}, and even downscaling~\cite{Oztireli15}.
To the best of our knowledge, however, SSIM-based indexes have never been adopted to train neural networks.

Recently, novel image quality indexes based on the properties of the HVS showed improved performance when compared to SSIM and MS-SSIM~\cite{Zhang12}.
One of these is the Information Weigthed SSIM (IW-SSIM), a modification of MS-SSIM that also includes a weighting scheme proportional to the local image information~\cite{IWSSIM}. Another is the Visual Information Fidelity (VIF), which is based on the amount of shared information between the reference and distorted image~\cite{VIF}. The Gradient Magnitude Similarity Deviation (GMSD) is characterized by simplified math and performance similar to that of SSIM, but it requires computing the standard deviation over the whole image~\cite{GMSD}. 
Finally, the Feature Similarity Index (FSIM), leverages the perceptual importance of phase congruency, and measures the dissimilarity between two images based on local phase congruency and gradient magnitude~\cite{FSIM}. FSIM has also been extended to FSIM$_c$, which can be used with color images.
Despite the fact that they offer an improved accuracy in terms of image quality, the mathematical formulation of these indexes is generally more complex than SSIM and MS-SSIM, and possibly not differentiable, making their adoption for optimization procedures not immediate.

%% file: new_loss_layers.tex

The loss layer of a neural network compares the output of the network with the ground truth, i.e., processed and reference patches, respectively, for the case of image processing.

In our work, we investigate the impact of different loss function layers for image processing. Consider the case of a network that performs denoising and demosaicking jointly. The insets in Figure~\ref{fig:teaser} show a zoom-in of different patches for the image in Figure~\ref{fig:teaser}(a) as processed by a network trained with different loss functions (see Section~\ref{sec:results} for the network's description). A simple visual inspection is sufficient to appreciate the practical implications of the discussion on \ltwo (Section~\ref{sec:related}). 

Specifically, Figure~\ref{fig:teaser}(d) shows that in flat regions the network strongly attenuates the noise, but it produces visible splotchy artifacts. This is because \ltwo penalizes larger errors, but is more tolerant to small errors, regardless of the underlying structure in the image; the HVS, on the other hand, is more sensitive to luminance and color variations in texture-less regions~\cite{Winkler2004}. A few splotchy artifacts are still visible, though arguably less apparent, in textured regions, see Figure~\ref{fig:teaser}(k). The sharpness of edges, however, is well-preserved by \ltwo, as blurring them would result in a large error. Note that these splotchy artifacts have been systematically observed before in the context of image processing with neural networks~\cite{Burger12}, but they have not been attributed to the loss function. In Section~\ref{sec:convergence} we show that the quality achieved by using \ltwo is also dependent on its convergence properties.

In this section we propose the use of different error functions. We provide a motivation for the different loss functions and we show how to compute their derivatives, which are necessary for the backpropagation step. We also share our implementation of the different layers that can be readily used within Caffe.

For an error function $\mathcal{E}$, the loss for a patch $P$ can be written as
\begin{equation}\label{eq:generalLoss}
{\cal L}^{\mathcal{E}}(P) = \frac{1}{N}\sum_{p \in P} \mathcal{E}(p),
\end{equation}
where $N$ is the number of pixels $p$ in the patch.

\subsection{The \lone error}\label{sec:Lone}
As a first attempt to reduce the artifacts introduced by the \ltwo loss function, we want to train the exact same network using \lone instead of \ltwo.
The two losses weigh errors differently---\lone does not over-penalize larger errors---and, consequently, they may have different convergence properties.

Equation~\ref{eq:generalLoss} for \lone is simply:
\begin{equation}
{\cal L}^{\ell_1}(P) = \frac{1}{N}\sum_{p \in P} \left| x(p) - y(p) \right|,
\end{equation}
where $p$ is the index of the pixel and $P$ is the patch; $x(p)$ and $y(p)$ are the values of the pixels in the processed patch and the ground truth respectively. The derivatives for the backpropagation are also simple, since $\partial {\cal L}^{\text{\lone}}(p)/\partial q = 0, \forall q \ne p$.
Therefore, for each pixel $p$ in the patch,
\begin{equation}
\partial {\cal L}^{\ell_1}(P)/\partial x(p)  = \text{sign}\left(x(p) - y(p)\right).
\end{equation}
The derivative of ${\cal L}^{\ell_1}$ is not defined at $0$. However, if the error is $0$, we do not need to update the weights, so we use the convention that sign$(0)=0$.
Note that, although ${\cal L}^{\ell_1}(P)$ is a function of the patch as a whole, the derivatives are back-propagated for each pixel in the patch.
Somewhat unexpectedly, the network trained with \lone provides a significant improvement for several of the issues discussed above, see Figure~\ref{fig:teaser}(e) where the splotchy artifacts in the sky are removed. Section~\ref{sec:convergence} analyzes the reasons behind this behavior. Although \lone shows improved performance over \ltwo, however, its results are still sub-optimal (see for instance the artifacts in the sky and at the boundary of the building in Figure~\ref{fig:jpegTeaser}(c) and \ref{fig:jpegTeaser}(g)).

\subsection{SSIM}\label{sec:ssim}
If the goal is for the network to learn to produce visually pleasing images, it stands to reason that the error function should be perceptually motivated, as is the case with SSIM.

SSIM for pixel $p$ is defined as
\begin{eqnarray}
\text{SSIM}(p) &=& \frac{2\mu_x\mu_y+C_1}{\mu_x^2+\mu_y^2+C_1} \cdot \frac{2 \sigma_{xy}+C_2}{\sigma_x^2+\sigma_y^2+C_2}\label{eq:origSSIM}\\
&=& l(p) \cdot cs(p)\label{eq:SSIM}
\end{eqnarray}
where we omitted the dependence of means and standard deviations on pixel $p$. Means and standard deviations are computed with a Gaussian filter with standard deviation $\sigma_G$, $G_{\sigma_G}$.
The loss function for SSIM can be then written setting $\mathcal{E} \left(p \right) = 1 - \text{SSIM}\left( p \right)$:
\begin{equation}\label{eq:FullSSIMloss}
{\cal L}^{\text{SSIM}}(P) = \frac{1}{N}\sum_{p \in P} 1 - \text{SSIM}(p).
\end{equation}
Equation~\ref{eq:origSSIM} highlights the fact that the computation of SSIM$(p)$ requires looking at a neighborhood of pixel $p$ as large as the support of $G_{\sigma_G}$. This means that ${\cal L}^{\text{SSIM}}(P)$, as well as its derivatives, cannot be calculated in some boundary region of $P$. This is not true for \lone or \ltwo, which only need the value of the processed and reference patch at pixel $p$.

However, the convolutional nature of the network allows us to write the loss as
\begin{equation}\label{eq:approxSSIMloss}
{\cal L}^{\text{SSIM}}(P) = 1 -  \text{SSIM}(\tilde{p}),
\end{equation}
where $\tilde{p}$ is the center pixel of patch $P$. Again, this is because, even though the network learns the weights maximizing SSIM for the central pixel, the learned kernels are then applied to all the pixels in the image. Note that the error can still be back-propagated to all the pixels within the support of $G_{\sigma_G}$ as they contribute to the computation of Equation~\ref{eq:approxSSIMloss}.

Recall that we have to compute the derivatives at $\tilde{p}$ with respect to any other pixel $q$ in patch $P$. More formally:
\begin{equation}
\begin{split}
\frac{\partial {\cal L}^{\text{SSIM}}(P)}{\partial x(q)} &= -\frac{\partial}{\partial x(q)} \text{SSIM}(\tilde{p}) \\
&= - \left(\frac{\partial l(\tilde{p})}{\partial x(q)}\cdot cs(\tilde{p}) + l(\tilde{p}) \cdot \frac{\partial cs(\tilde{p})}{\partial x(q)}\right),
\end{split}
\end{equation}
where $l(\tilde{p})$ and $cs(\tilde{p})$ are the first and second term of SSIM (Equation~\ref{eq:SSIM}) and their derivatives are
\begin{equation}\label{eq:dldx}
\frac{\partial l(\tilde{p})}{\partial x(q)} = 2\cdot G_{\sigma_G}(q-\tilde{p}) \cdot \left( \frac{\mu_y-\mu_x\cdot l(\tilde{p})}{\mu_x^2+\mu_y^2+C_1} \right)
\end{equation}
and
\begin{equation}\label{eq:dcsdx}
\begin{split}
\frac{\partial cs(\tilde{p})}{\partial x(q)} =& \frac{2}{\sigma_x^2+\sigma_y^2+C_2} \cdot G_{\sigma_G} (q-\tilde{p}) 
\cdot \big[ \left( y(q) - \mu_y\right) \\ 
&- cs(\tilde{p}) \cdot \left( x(q) - \mu_x\right) \big],
\end{split}
\end{equation}
where $G_{\sigma_G} (q-\tilde{p})$ is the Gaussian coefficient associated with pixel $q$.
We refer the reader to the additional material for the full derivation.

\subsection{MS-SSIM}\label{sec:ms-ssim}

The choice of $\sigma_G$ has an impact on the quality of the processed results of a network that is trained with SSIM, as can be seen from the derivatives in the previous section. Specifically, for smaller values of $\sigma_G$ the network loses the ability to preserve the local structure and the splotchy artifacts are reintroduced in flat regions, see Figure~\ref{fig:SSIM_sigmas}(e). For large values of $\sigma_G$, we observe that the network tends to preserve noise in the proximity of edges, Figure~\ref{fig:SSIM_sigmas}(c). See Section~\ref{sec:SSIMperf} for more details.

Rather than fine-tuning the $\sigma_G$, we propose to use the multi-scale version of SSIM, MS-SSIM. Given a dyadic pyramid of $M$ levels, MS-SSIM is defined as
\begin{equation}\label{eq:msssim}
\text{MS-SSIM}(p) = l^\alpha_M(p) \cdot \prod_{j = 1}^M cs^{\beta_j}_j(p)
\end{equation}
where $l_M$ and $cs_j$ are the terms we defined in Section~\ref{sec:ssim} at scale $M$ and $j$, respectively. For convenience, we set $\alpha = \beta_j = 1$, for $j = \{1,...,M\}$.  Similarly to Equation~\ref{eq:approxSSIMloss}, we can approximate the loss for patch $P$ with the loss computed at its center pixel $\tilde{p}$:
\begin{equation}\label{eq:MSSSIMloss}
{\cal L}^{\text{MS-SSIM}}(P) = 1 -  \text{MS-SSIM}(\tilde{p}).
\end{equation}
Because we set all the exponents of Equation~\ref{eq:msssim} to one, the derivatives of the loss function based on MS-SSIM can be written as
\begin{equation}
\begin{split}
&\frac{\partial {\cal L}^{\text{MS-SSIM}}(P)}{\partial x(q)}  \\
&= \left( \frac{\partial l_M(\tilde{p})}{\partial x(q)} + l_M(\tilde{p}) \cdot \sum_{i=0}^M\frac{1}{cs_i(\tilde{p})}\frac{\partial cs_i(\tilde{p})}{\partial x(q)}\right) \cdot \prod_{j=1}^M cs_j(\tilde{p}),
\end{split}
\end{equation}
where the derivatives of $l_j$ and $cs_j$ are the same as in Section~\ref{sec:ssim}. For the full derivation we refer the reader to the supplementary material.

Using ${\cal L}^{\text{MS-SSIM}}$ to train the network, Equation~\ref{eq:msssim} requires that we compute a pyramid of $M$ levels of patch $P$, which is a computationally expensive operation given that it needs to be performed at each iteration.  To avoid this, we propose to approximate and replace the construction of the pyramid: instead of computing $M$ levels of the pyramid, we use $M$ different values for $\sigma_G$, each one being half of the previous, on the full-resolution patch. Specifically, we use $\sigma_G^i = \{0.5, 1, 2, 4, 8\}$ and define $cs_i \triangleq G_{\sigma_G^i}\cdot cs_0(\tilde{p})$ and $\partial cs_i(\tilde{p})/\partial x(q) \triangleq G_{\sigma_G^i} \cdot \partial cs_0(\tilde{p})/\partial x(q)$, where the Gaussian filters are centered at pixel $\tilde{p}$, and ``$\cdot$'' is a point-wise multiplication. The terms depending on $l_M$ can be defined in a similar way. We use this trick to speed up the training in all of our experiments.

\subsection{The best of both worlds: MS-SSIM + \lone}
By design, both MS-SSIM and SSIM are not particularly sensitive to uniform biases (see Section~\ref{sec:SSIMperf}). This can cause changes of brightness or shifts of colors, which typically become more dull. However, MS-SSIM preserves the contrast in high-frequency regions better than the other loss functions we experimented with.
On the other hand, \lone preserves colors and luminance---an error is weighed equally regardless of the local structure---but does not produce quite the same contrast as MS-SSIM.

To capture the best characteristics of both error functions, we propose to combine them:
\begin{equation}\label{eq:ourLoss}
{\cal L}^{\text{Mix}} = \alpha \cdot {\cal L}^{\text{MS-SSIM}} + (1-\alpha) \cdot G_{\sigma_G^M} \cdot {\cal L}^{\text{\lone}},
\end{equation}
where we omitted the dependence on patch $P$ for all loss functions, and we empirically set $\alpha = 0.84$.\footnote{We chose the value for this parameter so that the contribution of the two losses would be roughly balanced. While we tested a few different values, we did not perform a thorough investigation. We did, however, notice that the results were not significantly sensitive to small variations of $\alpha$.} The derivatives of Equation~\ref{eq:ourLoss} are simply the weighed sum of the derivatives of its two terms, which we compute in the previous sections. Note that we add a point-wise multiplication between $G_{\sigma_G^M}$ and ${\cal L}^{\text{\lone}}$: this is because MS-SSIM propagates the error at pixel $q$ based on its contribution to MS-SSIM of the central pixel $\tilde{p}$, as determined by the Gaussian weights, see Equations~\ref{eq:dldx} and~\ref{eq:dcsdx}.

%% file: results.tex
For our analysis of the different loss functions we focus on joint demosaicking plus\ denoising, a fundamental problem in image processing. We also confirm our findings by testing the different loss functions on the problems of super-resolution and JPEG artifacts removal.

 \begin{figure*}
 	\centering
 	\subfloat[LR]{\includegraphics[width=.237\columnwidth]{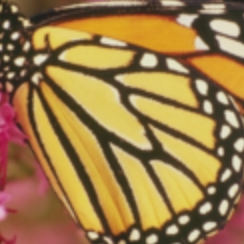}}\
 	\subfloat[\ltwo]{\includegraphics[width=.237\columnwidth]{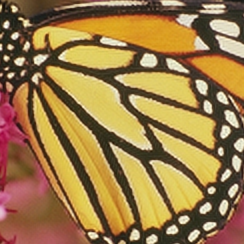}}\
 	\subfloat[\lone]{\includegraphics[width=.237\columnwidth]{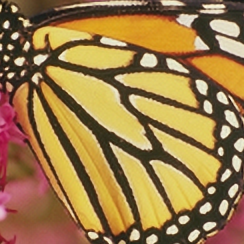}}\
 	\subfloat[Mix]{\includegraphics[width=.237\columnwidth]{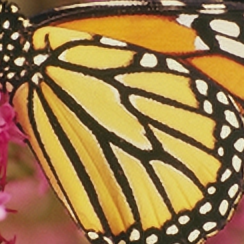}}\
 	\subfloat[LR]{\includegraphics[width=0.237\columnwidth]{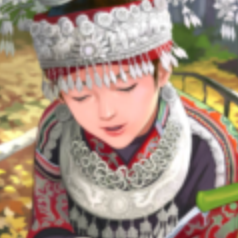}}\
 	\subfloat[\ltwo]{\includegraphics[width=0.237\columnwidth]{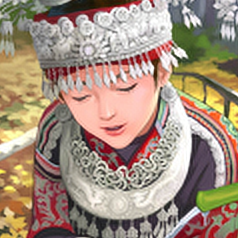}}\
 	\subfloat[\lone]{\includegraphics[width=0.237\columnwidth]{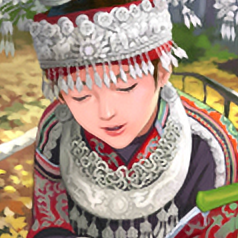}}\
 	\subfloat[Mix]{\includegraphics[width=0.237\columnwidth]{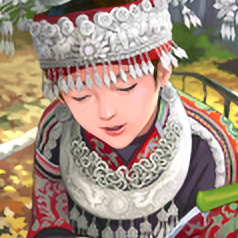}}
 	\caption{\small{Results for super-resolution. Notice the grating artifacts on the black stripes of the wing and around the face of the girl produced by \ltwo.}}\label{fig:SR}
 \end{figure*}
 
 \begin{figure*}
 	\centering
 	\subfloat[JPEG]{\includegraphics[width=.237\columnwidth]{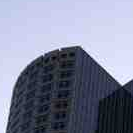}}\
 	\subfloat[\ltwo]{\includegraphics[width=.237\columnwidth]{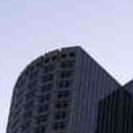}}\
 	\subfloat[\lone]{\includegraphics[width=.237\columnwidth]{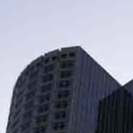}}\
 	\subfloat[Mix]{\includegraphics[width=.237\columnwidth]{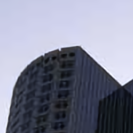}}\
 	\subfloat[JPEG]{\includegraphics[width=0.237\columnwidth]{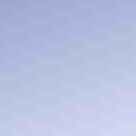}}\
 	\subfloat[\ltwo]{\includegraphics[width=0.237\columnwidth]{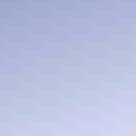}}\
 	\subfloat[\lone]{\includegraphics[width=0.237\columnwidth]{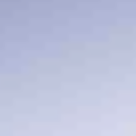}}\
 	\subfloat[Mix]{\includegraphics[width=0.237\columnwidth]{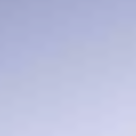}}
 	\caption{\small{Results for JPEG de-blocking. The insets are taken from the image in Figure~\ref{fig:teaser}. Notice the artifacts around the edges (a)-(c) and how Mix (d) removes them better than either \lone or \ltwo. Mix also outperforms the other metrics in the relatively flat regions, where the blocketization is more apparent, e.g., (e)-(h).}}\label{fig:jpegTeaser}
 \end{figure*}

\subsection{Joint denoising and demosaicking}\label{sec:resultDemosaic}
We define a fully convolutional neural network (CNN) that takes a 31$\times$31$\times$3 input. The first layer is a 64$\times$9$\times$9$\times$3 convolutional layer, where the first term indicates the number of filters and the remaining terms indicate their dimensions. The second convolutional layer is 64$\times$5$\times$5$\times$64, and the output layer is a 3$\times$5$\times$5$\times$64 convolutional layer. We apply parametric rectified linear unit (PReLU) layers to the output of the inner convolutional layers~\cite{He15}. The input to our network is obtained by bilinear interpolation of a 31$\times$31 Bayer patch, which results in a 31$\times$31$\times$3 RGB patch; in this sense the network is really doing joint denoising + demosaicking and super-resolution.
We trained the network considering different cost functions (\lone, \ltwo, SSIM$_5$, SSSIM$_9$, MS-SSIM and MS-SSIM+\lone)\footnote{SSIM$_k$ means SSIM computed with $\sigma_G=k$.} on a training set of 700 RGB images taken from the MIT-Adobe FiveK Dataset~\cite{fivek}, resized to a size of 999$\times$666. To simulate a realistic image acquisition process, we corrupted each image with a mix of photon shot and zero-mean Gaussian noise, and introduced clipping due to the sensor zero level and saturation. We used the model proposed by Foi et al.~\cite{Foi09} for this task, with parameters $a=0.005$ and $b=0.0001$. The average PSNR for the testing images after adding noise was 28.24dB, as reported in Table~\ref{tab:tables}. Figure~\ref{fig:teaser}(c) shows a typical patch corrupted by noise. We used 40 images from the same dataset for testing (the network did not see this subset during training).

Beyond considering different cost functions for training, we also compare the output of our network with the images obtained by a state-of-the-art denoising method, BM3D, operating directly in the Bayer domain~\cite{Danielyan09}, followed by a standard demosaicking algorithm~\cite{Zhang05}. Since BM3D is designed to deal with Gaussian noise, rather than the more realistic noise model we use, we apply a Variance Stabilizing Transform~\cite{Foi09} to the image data in Bayer domain before applying BM3D, and its inverse after denoising. This is exactly the strategy suggested by the authors of BM3D for RAW data~\cite{Danielyan09}, the paper we compare against for BM3D.

Figure~\ref{fig:teaser} and~\ref{fig:denoise} show several results and comparisons between the different networks. Note the splotchy artifacts for the \ltwo network on flat regions, and the noise around the edges for the SSIM$_5$ and SSIM$_9$ networks. The network trained with MS-SSIM addresses these problems but tends to render the colors more dull, see Section~\ref{sec:SSIMperf} for an explanation. The network trained with MS-SSIM+\lone generates the best results. Some differences are difficult to see in side-by-side comparisons, please refer to the additional material, which allows to flip between images.

We also perform a quantitative analysis of the results. We evaluate several image quality metrics on the output of the CNNs trained with the different cost functions and with BM3D~\cite{Danielyan09}, which is the state of the art in denoising. The image quality indexes, range from the traditional \ltwo metric and PSNR, to the most refined, perceptually inspired metrics, like FSIM~\cite{FSIM}.
The average values of these metrics on the testing dataset are reported in Table~\ref{tab:tables}.
When the network is trained using \lone as a cost function, instead of the traditional \ltwo, the average quality of the output images is superior for all the quality metrics considered. It is quite remarkable to notice that, even when the \ltwo or PSNR metrics are used to evaluate image quality, the network trained with the \lone loss outperforms the one trained with the \ltwo loss. We offer an explanation of this in Section~\ref{sec:convergence}. On the other hand, we note that the network trained with SSIM performs either at par or slightly worse than the one trained with \lone, both on traditional metrics and on perceptually-inspired losses. The network trained on MS-SSIM performs better than the one based on SSIM, but still does not clearly outperforms \lone. This is due to the color shifts mentioned above and discussed in Section~\ref{sec:SSIMperf}. However, the network that combines MS-SSIM and \lone achieves the best results on all of the image quality metrics we consider.

For this image restoration task, we also evaluated the combination of \ltwo and MS-SSIM to test whether the use of \lone is critical to the performance of Mix. The results on all the indexes considered in Table~\ref{tab:tables} show that Mix is still superior. As expected, however, \ltwo+ MS-SSIM outperforms \ltwo alone on the perceptual metrics (see supplementary material).

\subsection{Super-resolution}\label{sec:resultSuperRez}
We also verify the outcome of our analysis on the network for super-resolution proposed by Dong et al.~\cite{Fleet14}. We start from the network architecture they propose, and make a few minor but important changes to their approach. First we use PReLU, instead of ReLU, layers. Second we use bilinear instead of bicubic interpolation for initialization (we do this for both training and testing). The latter introduces high-frequency artifacts that hurt the learning process. Finally we train directly on the RGB data. We made these changes for all the loss functions we test, including \ltwo, which also serves as a comparison with the work by Dong et al. We use this modified network in place of their proposed architecture to isolate the contribution of the loss layer to the results---and because it produces better or equal results than the one produced by the original architecture. Figure~\ref{fig:SR} shows some sample results. Given the results of the previous section, we only compared the results of \lone, \ltwo, MS-SSIM, and Mix, see Table~\ref{tab:tables}. An analysis of the table brings similar considerations as for the case of joint denoising and demosaicking.

\subsection{JPEG artifacts removal}\label{sec:resultJPEG}
Our final benchmark is JPEG artifact removal. We use the same network architecture and ground truth data as for joint denoising and demosaicking, but we create the corrupted data by means of aggressive JPEG compression. For this purpose we use the Matlab's function \emph{imwrite} with a quality setting of 25. Note that the corruption induced by JPEG compression is spatially variant and has a very different statistic than the noise introduced in Section~\ref{sec:resultDemosaic}. We generate the input patches with two different strides. First, we use a stride of $8$, which causes the boundaries of the $8\times 8$ DCT blocks to be aligned in each 31$\times$31 patch. We also use a stride of $7$, which causes the grid of blocks to be in different locations in each patch.
We found that the latter strategy better removes the JPEG artifacts while producing sharper images, which is why we ran all of our experiments with that configuration. Again, we only compared the results of \lone, \ltwo, MS-SSIM, and Mix, see Table~\ref{tab:tables}. Figure~\ref{fig:jpegTeaser} shows that our loss function, Mix, outperforms \lone and \ltwo on uniform regions and that it attenuates the ringing artifacts around the edge of the building better. More results are shown in Figure~\ref{fig:JPEG}.

More results and comparisons, both numerical and visual, can be found in the supplementary material.

\begin{figure*}
\centering
\includegraphics[width=.245\textwidth]{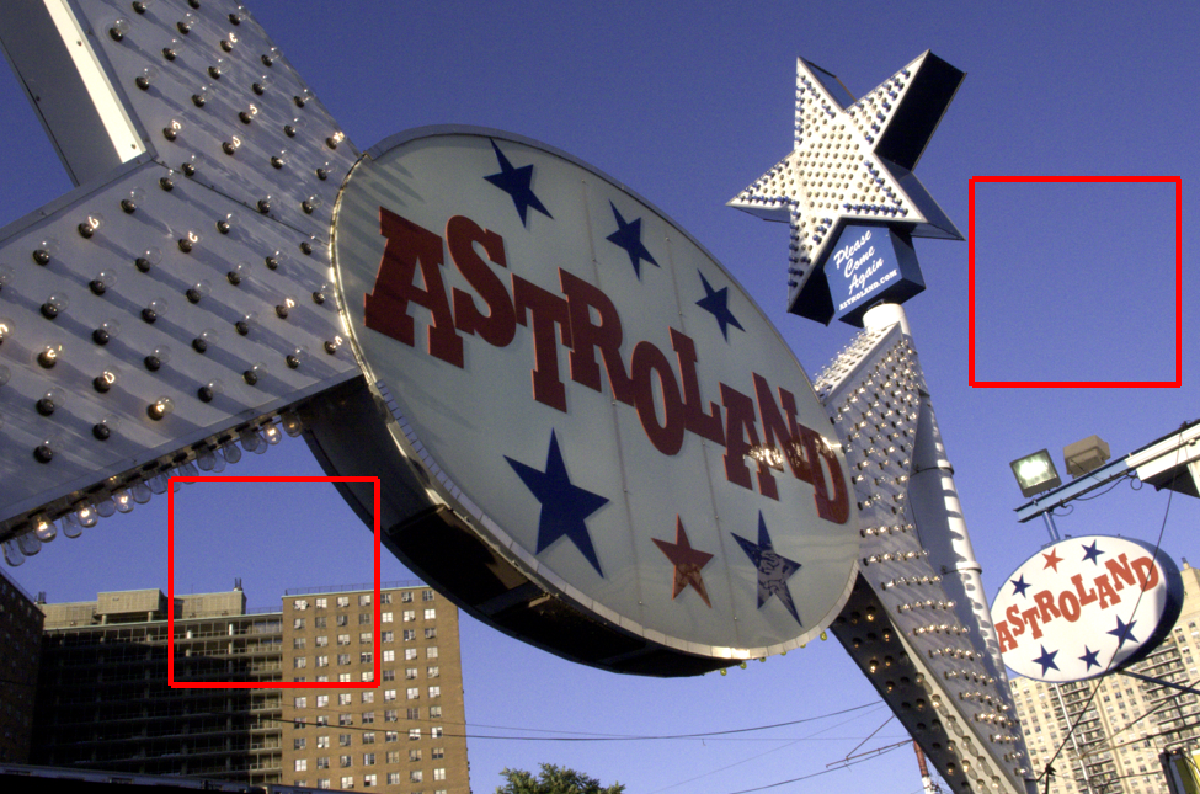}
\includegraphics[width=.245\textwidth]{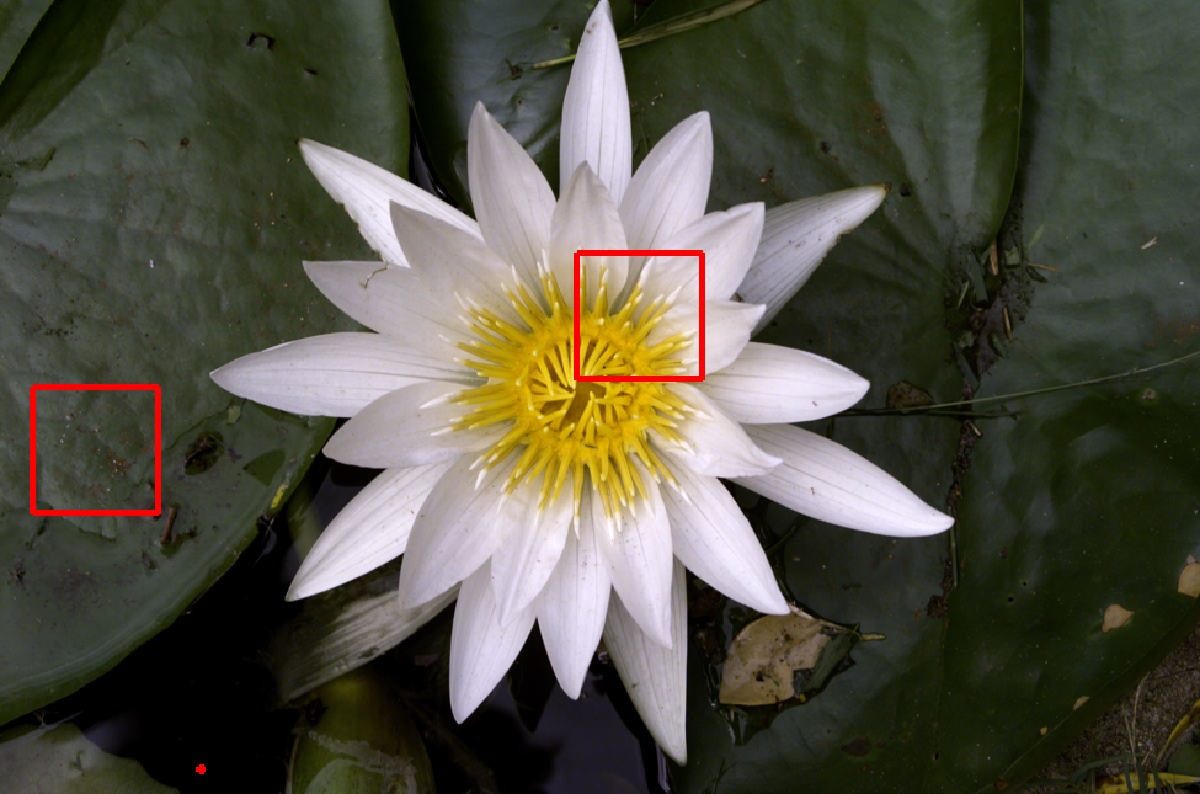}
\includegraphics[trim={0 1cm 0 0}, clip, width=.245\textwidth]{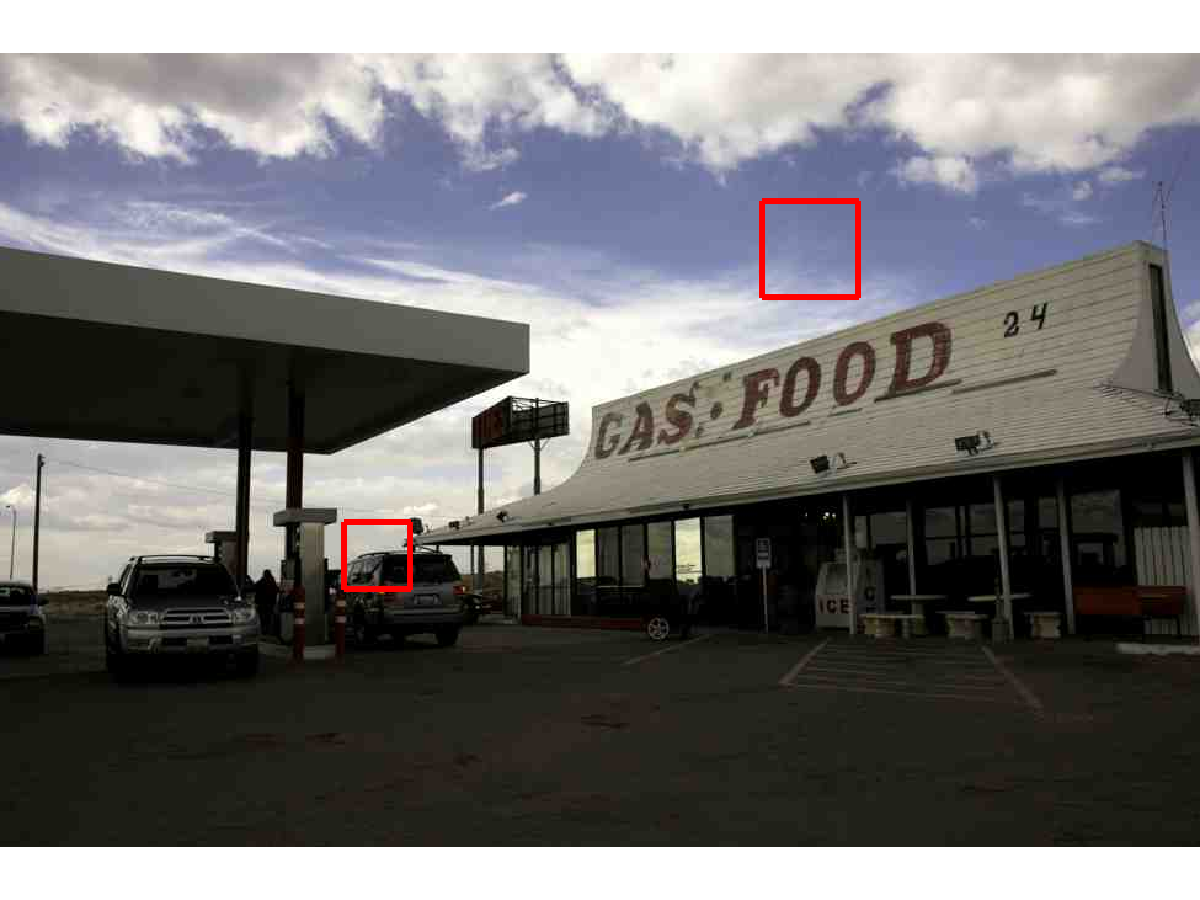}
\includegraphics[trim={0 1cm 0 0}, clip, width=.245\textwidth]{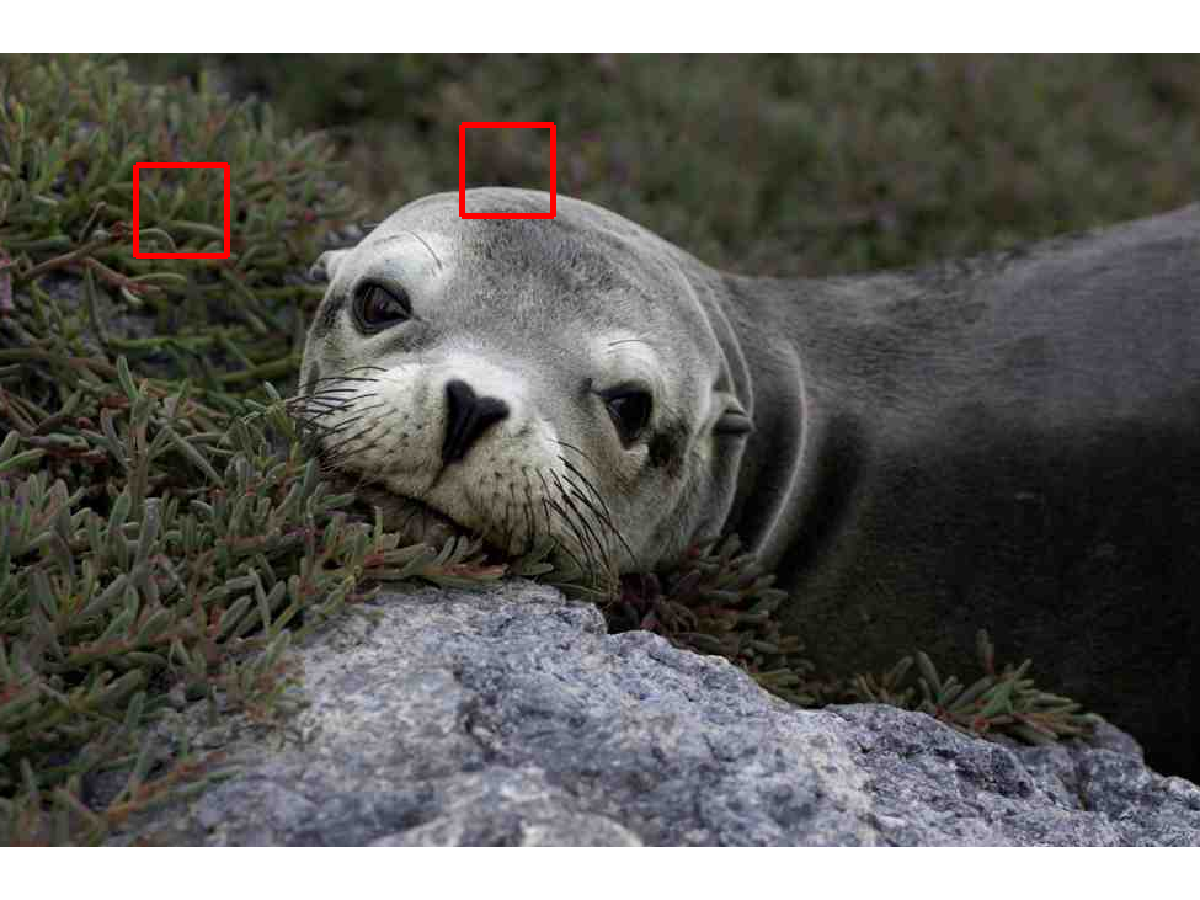}
\caption{\small{The reference images for the details shown in Figures~\ref{fig:denoise} and~\ref{fig:JPEG}.}}\label{fig:reference}
		
\includegraphics[trim={1.2cm 13.5cm 1.2cm 1cm}, clip, width=\textwidth]{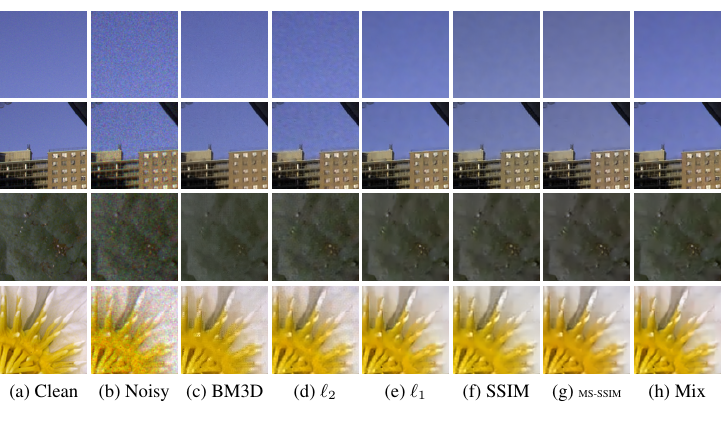}
\caption{\small{Results for denosing+demosaicking for different approaches. The noisy patches are obtained by simple bilinear interpolation. Note the splotchy artifacts \ltwo produces in flat regions. Also note the change in colors for SSIM-based losses. The proposed metric, MS-SSIM+\lone referred to as Mix, addresses the former issues. Reference images are shown in Figure~\ref{fig:reference}.}}\label{fig:denoise}

\includegraphics[trim={1.2cm 6.5cm 1.2cm 11cm}, clip, width=\textwidth]{results_pdf}
\caption{\small{Results for JPEG de-blocking for different approaches. Note that Mix outperforms both \ltwo and \lone on uniform regions as well as on the ringing artifacts at the in high-gradient regions. Reference images are shown in Figure~\ref{fig:reference}.}}\label{fig:JPEG}
\end{figure*}

\input{tables2}

%% file: tables2.tex

\begin{table*}
\footnotesize
\begin{center}
\renewcommand{\arraystretch}{0.9}
\begin{tabular}{|c|c|c|c|c|c|c|c|c|}\hline
\bf{Denoising + demosaicking} & & & \multicolumn{6}{|c|}{Training cost function}\\
\hline
Image quality metric & Noisy & $BM3D$ & \ltwo & \lone & SSIM$_5$ & SSIM$_9$ & MS-SSIM & Mix \\
\hline
1000 $\cdot$ \ltwo & 1.65 & 0.45 & 0.56 & 0.43 & 0.58 & 0.61 & 0.55 & \bf{0.41}\\
PSNR & 28.24 & 34.05 & 33.18 & 34.42 & 33.15 & 32.98 & 33.29 & \bf{34.61}\\
1000 $\cdot$ \lone & 27.36 & 14.14 & 15.90 & 13.47 & 15.90 & 16.33 & 15.99 & \bf{13.19}\\
SSIM & 0.8075 & 0.9479 & 0.9346 & 0.9535 & 0.9500 & 0.9495 & 0.9536 & \bf{0.9564}\\
MS-SSIM & 0.8965 & 0.9719 & 0.9636 & 0.9745 & 0.9721 & 0.9718 & 0.9741 & \bf{0.9757}\\
IW-SSIM & 0.8673 & 0.9597 & 0.9473 & 0.9619 & 0.9587 & 0.9582 & 0.9617 & \bf{0.9636}\\
GMSD & 0.1229 & 0.0441 & 0.0490 & 0.0434 & 0.0452 & 0.0467 & 0.0437 & \bf{0.0401}\\
FSIM & 0.9439 & 0.9744 & 0.9716 & 0.9775 & 0.9764 & 0.9759 & 0.9782 & \bf{0.9795}\\
FSIM$_c$ & 0.9381 & 0.9737 & 0.9706 & 0.9767 & 0.9752 & 0.9746 & 0.9769 & \bf{0.9788}\\
\hline
\end{tabular}


\vspace{2mm}
\renewcommand{\arraystretch}{0.9}
\begin{tabular}{|c|c|c|c|c|c|}
\hline
\bf{Super-resolution} & & \multicolumn{4}{|c|}{Training cost function}\\
\hline
Image quality metric  & Bilinear & $\ell_2$ & $\ell_1$ & MS-SSIM & Mix \\
\hline
$1000 \cdot \ell_2$ & 2.5697 & 1.2407 & 1.1062 & 1.3223 & \bf{1.0990}\\
PSNR & 27.16 & 30.66 & 31.26 & 30.11 & \bf{31.34}\\
$1000 \cdot \ell_1$ & 28.7764 & 20.4730 & 19.0643 & 22.3968 & \bf{18.8983}\\
SSIM & 0.8632 & 0.9274 & 0.9322 & 0.9290 & \bf{0.9334}\\
MS-SSIM & 0.9603 & 0.9816 & 0.9826 & 0.9817 & \bf{0.9829}\\
IW-SSIM & 0.9532 & 0.9868 & 0.9879 & 0.9866 & \bf{0.9881}\\
GMSD & 0.0714 & 0.0298 & 0.0259 & 0.0316 & \bf{0.0255}\\
FSIM & 0.9070 & 0.9600 & 0.9671 & 0.9601 & \bf{0.9680}\\
FSIM$_c$ & 0.9064 & 0.9596 & 0.9667 & 0.9597 & \bf{0.9677}\\
\hline
\end{tabular}
\renewcommand{\arraystretch}{1}

\vspace{2mm}
\renewcommand{\arraystretch}{0.9}
\begin{tabular}{|c|c|c|c|c|c|}
\hline
\bf{JPEG de-blocking} & & \multicolumn{4}{|c|}{Training cost function}\\
\hline
Image quality metric  & Original JPEG & $\ell_2$ & $\ell_1$ & MS-SSIM & Mix \\
\hline
$1000 \cdot \ell_2$ & 0.6463 & 0.6511 & 0.6027 & 1.9262 & \bf{0.5580}\\
PSNR & 32.60 & 32.73 & 32.96 & 27.66 & \bf{33.25}\\
$1000 \cdot \ell_1$ & 16.5129 & 16.2633 & 16.0687 & 33.6134 & \bf{15.5489}\\
SSIM & 0.9410 & 0.9427 & 0.9467 & 0.9364 & \bf{0.9501}\\
MS-SSIM & 0.9672 & 0.9692 & 0.9714 & 0.9674 & \bf{0.9734}\\
IW-SSIM & 0.9527 & 0.9562 & 0.9591 & 0.9550 & \bf{0.9625}\\
GMSD & 0.0467 & 0.0427 & 0.0413 & 0.0468 & \bf{0.0402}\\
FSIM & 0.9805 & 0.9803 & 0.9825 & 0.9789 & \bf{0.9830}\\
FSIM$_c$ & 0.9791 & 0.9790 & 0.9809 & 0.9705 & \bf{0.9815}\\
\hline
\end{tabular}
\renewcommand{\arraystretch}{1}
\vspace{2mm}
\caption{\small{Average value of different image quality metrics on the testing dataset for the different cost functions. For SSIM, MS-SSIM, IW-SSIM, GMSD and FSIM the value reported here has been obtained as an average of the three color channels. Best results are shown in bold.
(Lower is better for \lone, \ltwo, and GMSD, higher is better for the others.)}}\label{tab:tables}
\vspace{-2mm}
\end{center}
\end{table*}

%% file: discussion.tex

\section{Discussion}\label{sec:discussion}
In this section we delve into a deeper analysis of the results. Among other considerations, we look into the convergence properties of the different losses, and we offer an interpretation of the reasons some losses perform better than others.

\subsection{Convergence of the loss functions}\label{sec:convergence}

\begin{figure}
\begin{center}
\includegraphics[trim=0.5cm 9.5cm 1cm 9.5cm, clip,width=.9\columnwidth]{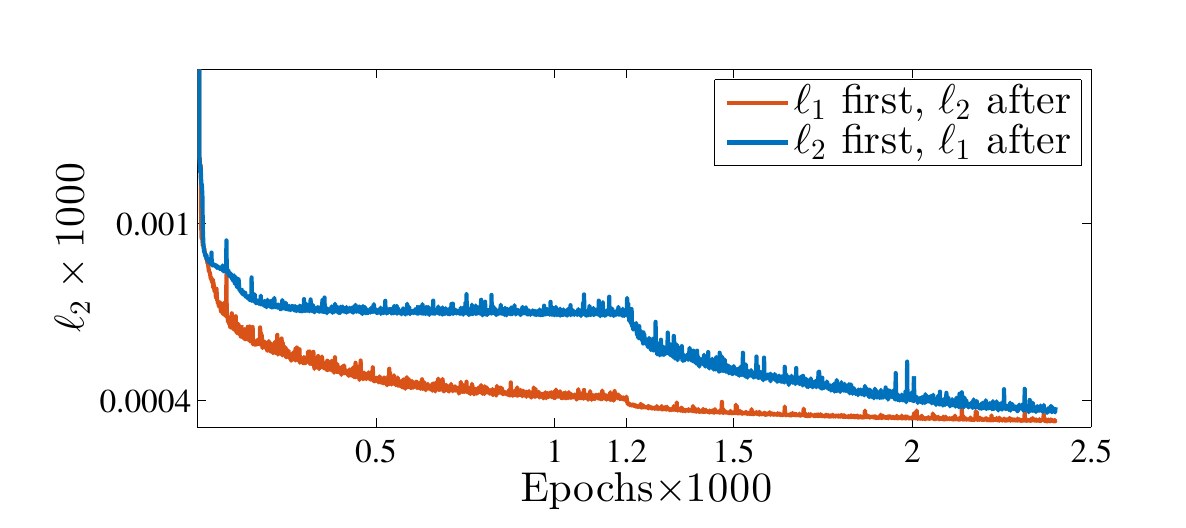}
\end{center}
\caption{\small{\ltwo loss on the testing set for the two networks that switch loss functions during training.}}\label{fig:convergencePlot}
\end{figure}

\begin{figure*}
\centering
\subfloat[\lone]{\includegraphics[width=.245\columnwidth]{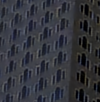}}\
\subfloat[\lone$\rightarrow$\ltwo]{\includegraphics[width=.245\columnwidth]{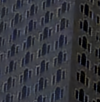}}\
\subfloat[\ltwo]{\includegraphics[width=.245\columnwidth]{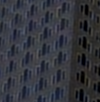}}\
\subfloat[\ltwo$\rightarrow$\lone]{\includegraphics[width=.245\columnwidth]{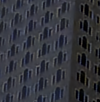}}\
\subfloat[\lone]{\includegraphics[width=.245\columnwidth]{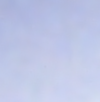}}\
\subfloat[\lone$\rightarrow$\ltwo]{\includegraphics[width=.245\columnwidth]{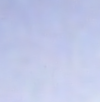}}\
\subfloat[\ltwo]{\includegraphics[width=.245\columnwidth]{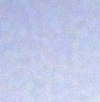}}\
\subfloat[\ltwo$\rightarrow$\lone]{\includegraphics[width=.245\columnwidth]{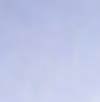}}
\caption{\small{Visual results of the networks trained alternating loss functions. The insets here correspond the regions marked in Figure~\ref{fig:teaser}. Note that $\ell_i\rightarrow\ell_j$ indicates a network trained first with $\ell_i$ and  then with $\ell_j$.}}\label{fig:convergenceVisual}
\end{figure*}

Table~\ref{tab:tables} highlights an unexpected result: even after convergence, CNNs trained on one loss function can outperform another network even based on the very loss with which the second was trained. Consider, for instance, the two networks trained with \lone and \ltwo respectively for joint denoising and demosaicking: the table shows that the network trained with \lone achieves a lower \ltwo loss than then network trained with \ltwo. Note that we ran the training multiple times with different initializations.
We hypothesize that this result may be related to the smoothness and the local convexity properties of the two measures: \ltwo gets stuck more easily in a local minimum, while for \lone it may be easier to reach better minimum, both in terms of \lone and \ltwo---the ``good'' minima of the two should be related, after all.
To test this hypothesis, we ran an experiment in which we take two networks trained with \lone and \ltwo respectively, and train them again until they converge using the other loss.
Figure \ref{fig:convergencePlot} shows the \ltwo loss computed on the testing
set at different training iterations for either network.
The network trained with \lone only (before epoch 1200 in the plot) achieves a better \ltwo loss than the one trained \mbox{with \ltwo}.
However, after switching the training loss functions, both networks yield a lower \ltwo loss, confirming that the \ltwo network was previously stuck in a local minimum. 
While the two networks achieve a similar \ltwo loss, they converge to different regions of the space of parameters. At visual inspection the network trained with \ltwo first and \lone after produces results similar to those of \lone alone; the output of the network trained with \lone first and \ltwo after is still affected by splotchy artifacts in flat areas, though it is better than \ltwo alone, see Figure~\ref{fig:convergenceVisual}.
Specifically, the network trained with \lone first and \ltwo afterwards achieves an \ltwo loss of $0.3896\cdot10^3$, which is the lowest across all the networks we trained, confirming that the network trained with \ltwo alone had convergence issues. However, neither of these two networks outperforms Mix on any of the perceptual metrics we use, confirming that the advantage of the proposed loss goes beyond convergence considerations. Table~\ref{tab:l1l2} shows the complete evaluation of the two networks, where bold indicates a result that is better than any of the results shown in Table~\ref{tab:tables}.

\subsection{On the performance of SSIM and MS-SSIM}\label{sec:SSIMperf}
\begin{figure*}
	\centering
	\subfloat[Clean image]{\includegraphics[height=0.92in]{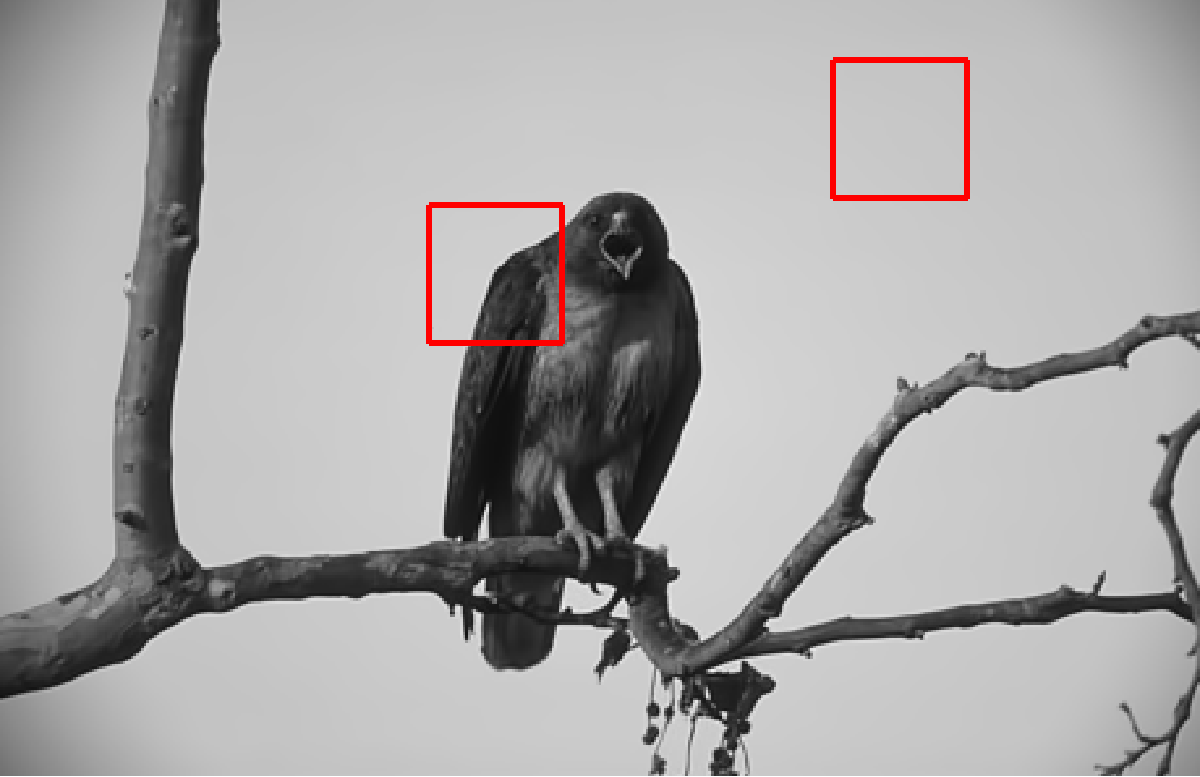}}\
	\subfloat[SSIM$_1$]{\includegraphics[height=0.92in]{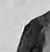}}\
	\subfloat[SSIM$_3$]{\includegraphics[height=0.92in]{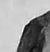}}\
	\subfloat[SSIM$_9$]{\includegraphics[height=0.92in]{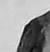}}\
	\subfloat[SSIM$_1$]{\includegraphics[height=0.92in]{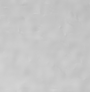}}\
	\subfloat[SSIM$_3$]{\includegraphics[height=0.92in]{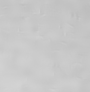}}\
	\subfloat[SSIM$_9$]{\includegraphics[height=0.92in]{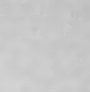}}
	\caption{Comparison of the results of networks trained with SSIM with different sigmas (SSIM$_k$ means $\sigma_G=k$). Insets (b)--(d), show an increasingly large halo of noise around the edge: smaller values of $\sigma$ help at edges. However, in mostly flat regions, larger values of $\sigma$ help reducing the splotchy artifacts (e)--(g). Best viewed by zooming in on the electronic copy.}\label{fig:SSIM_sigmas}
\end{figure*}

\begin{figure*}
	\subfloat[ ]{\includegraphics[width=0.49\columnwidth, trim=0 6.5cm 2.5cm 0, clip=true, scale=1]{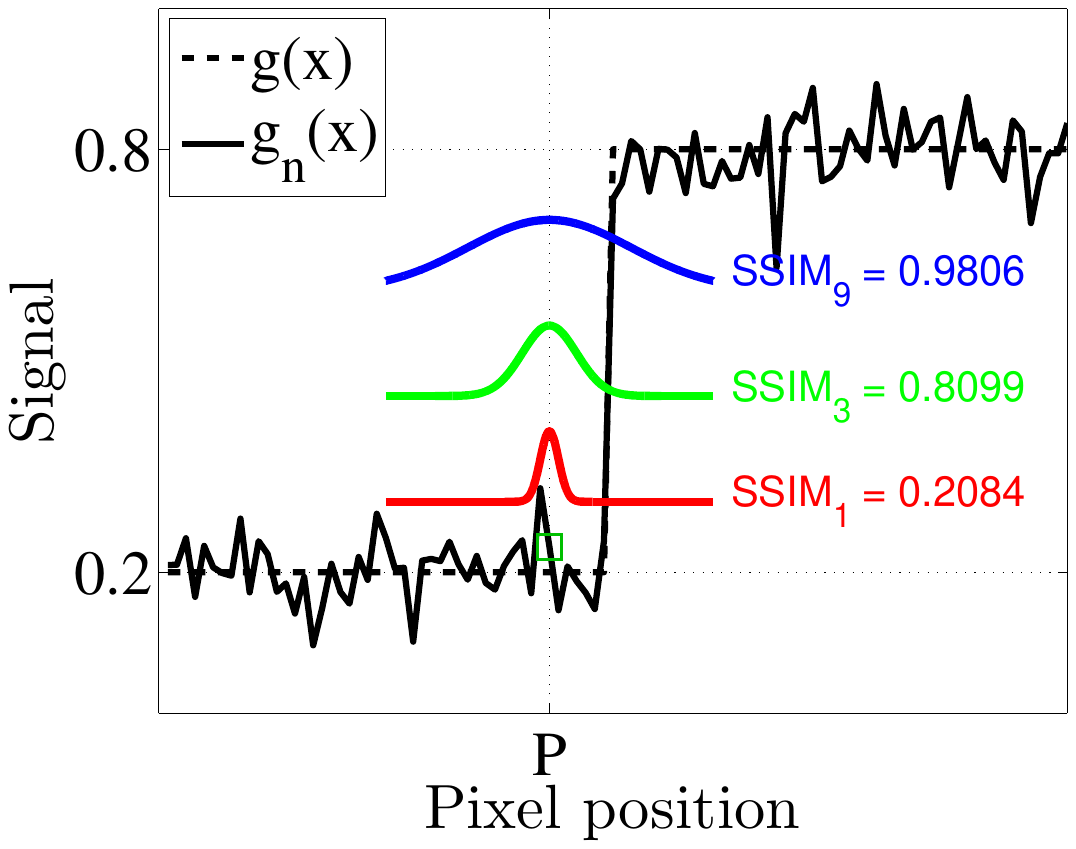}}\enskip
	\subfloat[ ]{\includegraphics[width=0.49\columnwidth, trim=0 6.5cm 2.5cm 0, clip=true, scale=1]{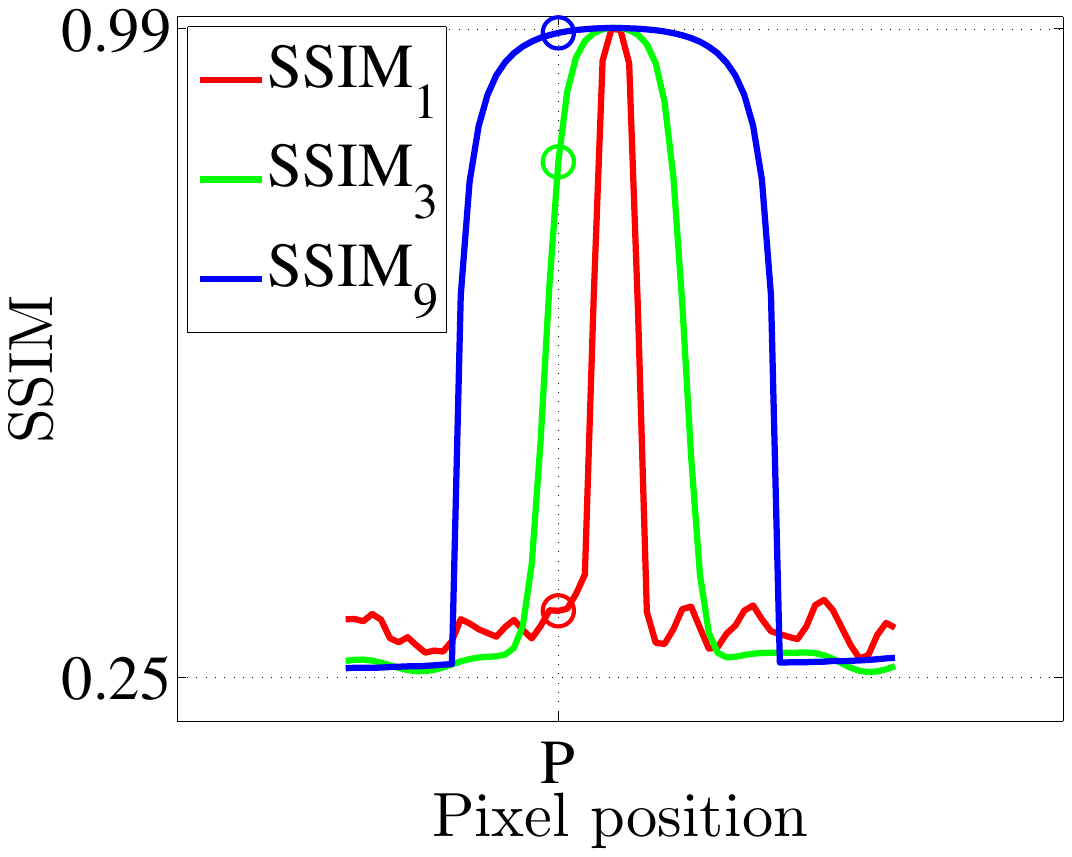}}\enskip
	\subfloat[ ]{\includegraphics[width=0.49\columnwidth, trim=0 6.5cm 2.5cm 0, clip=true, scale=1]{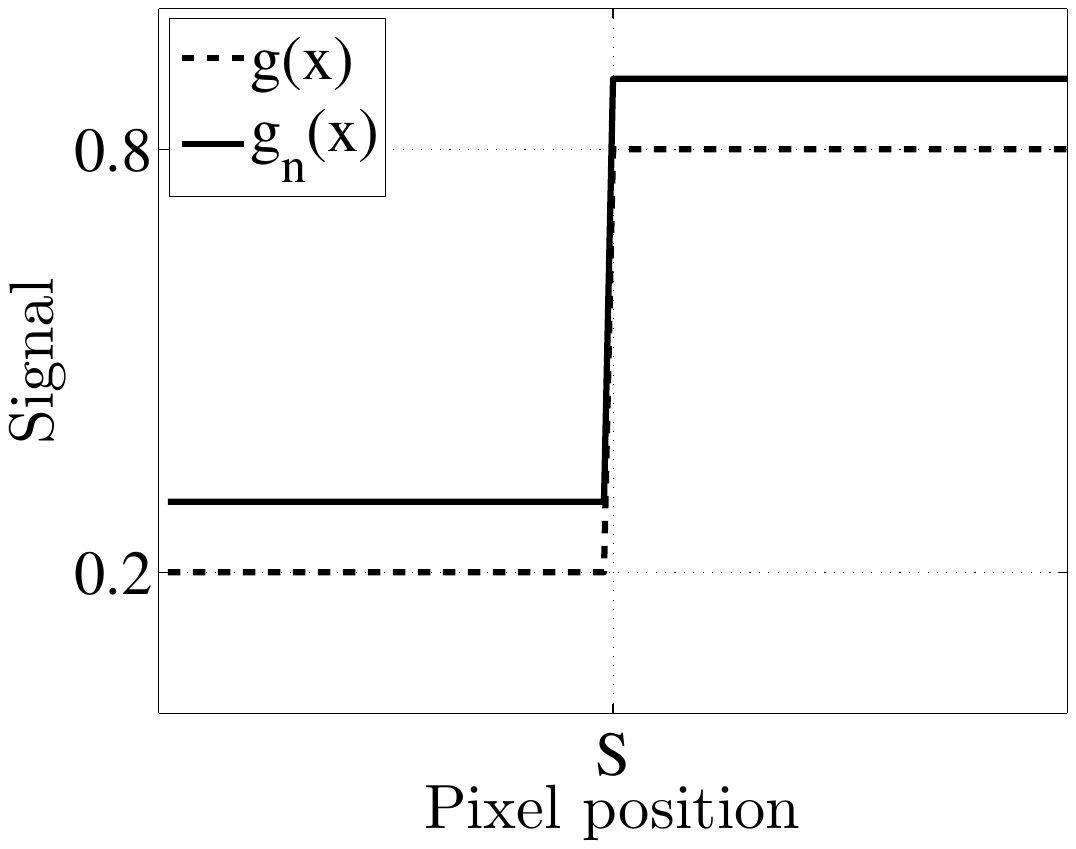}}\enskip
	\subfloat[ ]{\includegraphics[width=0.49\columnwidth, trim=0 6.5cm 2.5cm 0, clip=true, scale=1]{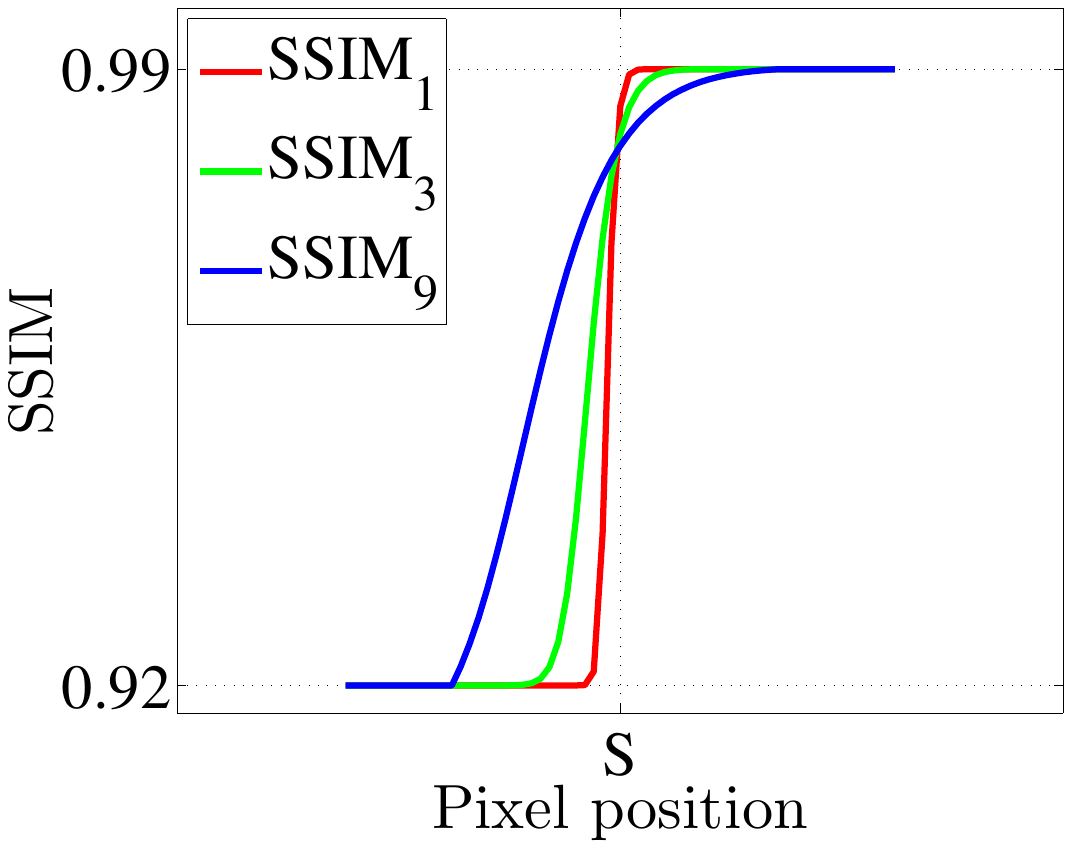}}
	\caption{\small{Panels (a) and (b) show how the size of the support affects SSIM for a pixel P close to an edge, in the presence of zero-mean Gaussian noise. Panels (c) and (d) show that SSIM is less sensitive to a uniform bias in bright regions, such as the one to the right of pixel S (see Section~\ref{sec:SSIMperf}).}}\label{fig:ssim}
\end{figure*}

Table~\ref{tab:tables} also reveals that SSIM and MS-SSIM do not perform as well as \lone. This does not have to do with the convergence properties of these losses. To investigate this, we trained several SSIM networks with different $\sigma_G$'s and found that smaller values of $\sigma_G$ produce better results at edges, but worse results in flat regions, while the opposite is true for larger values, see Figure~\ref{fig:SSIM_sigmas}. This can be understood by looking at Figure~\ref{fig:ssim}(a), where the size of the support for the computation of SSIM for the three values of $\sigma_G$ is shown for a pixel P close to an edge.
A larger $\sigma_G$ is more tolerant to the same amount of noise because it detects the presence of the edge, which is known to have a masking effect for the HVS: in this toy example, SSIM$_9$ yields a higher value for the same pixel because its support spans both sides of the edge. Figure~\ref{fig:ssim}(b) shows SSIM across the edge of the same profile of (a), and particularly at pixel P. Note that SSIM$_9$ estimates a higher quality in a larger region around the step.

Despite of the size of its support, however, SSIM is not particularly sensitive to a uniform bias on a flat region. This is particularly true in bright regions, as shown in Figure~\ref{fig:ssim}(d), which plots the value of SSIM for the noisy signal of Figure~\ref{fig:ssim}(c). The same bias on the two sides of pixel S impacts SSIM's quality assessment differently.
Because the term $l(p)$ in Equation~\ref{eq:SSIM} measures the error in terms of a contrast, it effectively reduces its impact when the background is bright.
This is why, thanks to its multi-scale nature, MS-SSIM solves the issue of noise around edges, but does not solve the problem of the change colors in flat areas, in particular when at least one channel is strong, as is the case with the sky.

\begin{table}
\footnotesize
\begin{center}
\renewcommand{\arraystretch}{0.9}
\begin{tabular}{|c|c|c|}\hline
&\multicolumn{2}{|c|}{Training cost function}\\
\hline
Image quality metric & \ltwo first, \lone after& \lone first, \ltwo after\\
\hline
1000 $\cdot$ \ltwo & 0.3939 & \bf{0.3896}\\
PSNR & 34.76 & \bf{34.77}\\
1000 $\cdot$ \lone & \bf{12.7932} & 12.8919\\
SSIM & 0.9544 & 0.9540 \\
MS-SSIM & 0.9753 & 0.9748 \\
IW-SSIM & 0.9634 & 0.9624 \\
GMSD & 0.0432 & 0.0405 \\
FSIM & 0.9777 & 0.9781\\
FSIM$_c$ & 0.9770 & 0.9774\\
\hline
\end{tabular}
\caption{Average value of different image quality metrics for the networks trained for denoising + demosaicking with alternating loss functions. Bold indicates that the network achieves a better score than any of the networks in Table~\ref{tab:tables}, see Section~\ref{sec:convergence}.}\label{tab:l1l2}
\end{center}
\end{table}

Note that, while these are important observations because they generalize to any optimization problem that uses SSIM or MS-SSIM, the definition of a novel image quality metric is beyond the scope of this paper. We limit ourselves to the analysis of their performance within the context of neural networks.

Finally, we would like to point out that using SSIM-based losses in the context of color images introduces an approximation, since these metrics were originally designed for grayscale images. A more principled approach would be to base the loss function on a metric defined  specifically for color images, such as FSIM$_c$. However, to the best of our knowledge, no such metric has been proposed that is also differentiable. We plan on addressing this problem in future work.

\subsection{Are we committing the inverse crime?}\label{sec:remarks}
To numerically evaluate and compare the results of the different loss functions, we need access to the ground truth. For the applications described in this paper, this is only possible if we generate synthetic data. For instance, for the super-resolution experiments, we generate the low-resolution inputs by low-passing and down-sampling the original images. However, when generating synthetic data, one needs to be careful about the validity of the results. In the context of inverse problems, Colton and Kress introduce the concept of \emph{inverse crime}, which is ``committed'' when synthetic data is fabricated with a forward model that is related to the inverse solver (\cite{colton2012inverse}, page 133), potentially hiding shortcomings of the approach. By this definition, it could be argued that we did commit the inverse crime, as we generate the synthetic data for both training and testing with the same process. However, two considerations support our choice. First, while in general the inverse crime is to be avoided, the forward model should resemble the physical system as closely as possible (\cite{colton2012inverse}, page 163). In the super-resolution example described above, for instance, performing down-sampling followed by low-passing mimics what the PSF of the lens and the sensor do to the incoming light.  Second, we are studying the effect of the loss layer when all other conditions are equal: if we change the forward model, the quality of the results may degrade, but we expect the relationship between losses to hold. This is exactly what we observed when we tested the networks trained with the different losses on data generated with a varying standard deviation for the Gaussian filter. Changing the standard deviation from 3 to 5 pixels, for instance, causes the average MS-SSIM yielded by the network trained on Mix to drop from 0.9829 to 0.9752. However, the networks trained with \lone and \ltwo experience a similar drop,  yielding an average MS-SSIM of 0.9744 and 0.9748 respectively. We observed a similar behavior for the other metrics. These considerations also apply to joint denoising and demosaicking, as well as JPEG de-blocking.

%% file: conclusions.tex

We focus on an aspect of neural networks that is usually overlooked in the context of image restoration: the loss layer. We propose several alternatives to \ltwo, which is the \emph{de facto} standard, and we also define a novel loss. We use the problems of joint denoising and demosaicking, super-resolution, and JPEG artifacts removal for our tests. We offer a thorough analysis of the results in terms of both traditional and perceptually-motivated metrics, and show that the network trained with the proposed loss outperforms other networks. We also notice that the poor performance of \ltwo is partially related to its convergence properties, and this can help improve results of other \ltwo-based approaches. Because the networks we use are fully convolutional, they are extremely efficient, as they do not require an aggregation step. Nevertheless, thanks to the loss we propose, our joint denoising and demosaicking network outperforms CFA-BM3D, a variant of BM3D tuned for denoising in Bayer domain, which is the state-of-the-art denoising algorithm. Table~\ref{tab:tables} shows the superiority of Mix over all the other losses. This can also be appreciated by visual inspection even for networks trained with \lone, the closes competitor, which produce stronger artifacts (see, for instance, Figure~\ref{fig:jpegTeaser}). We also make the implementation of the layers described in this paper available to the research community.